\documentclass[a4paper]{article}
\usepackage{arxiv}
\usepackage[natbibapa]{apacite}
\usepackage{graphicx}
\usepackage[utf8]{inputenc} 
\usepackage[T1]{fontenc}    
\usepackage{hyperref}       
\usepackage{url}            
\usepackage{booktabs}       
\usepackage{amsfonts}       
\usepackage{nicefrac}       
\usepackage{microtype}      
\usepackage{xcolor}
\usepackage{color,xspace,tabularx}

\title{How Personal is Machine Learning Personalization?}
\author{
  Travis Greene  \\
  Institute of Service Science\\
  National Tsing Hua University\\
  Hsinchu, Taiwan \\
  \texttt{travis.greene@iss.nthu.edu.tw
} \\
   \And
 Galit Shmueli \\
  Institute of Service Science\\
  National Tsing Hua University\\
  Hsinchu, Taiwan  \\
  \texttt{galit.shmueli@iss.nthu.edu.tw
} \\
}

\begin{document}
\maketitle

\begin{abstract}
Though used extensively, the concept and process of machine learning (ML) personalization have generally received little attention from academics, practitioners, and the general public. We describe the ML approach as relying on the \emph{metaphor of the person as a feature vector} and contrast this with humanistic views of the person. 
In light of the recent calls by the IEEE to consider the effects of ML on human well-being, we ask whether ML personalization can be reconciled with these humanistic views of the person, which highlight the importance of moral and social identity. As human behavior increasingly becomes digitized, analyzed, and predicted, to what extent do our subsequent decisions about what to choose, buy, or do, made both by us and others, reflect who we are as persons? This paper first explicates the term \textit{personalization} by considering ML personalization and highlights its relation to humanistic conceptions of the \textit{person}, then proposes several dimensions for evaluating the degree of personalization of ML personalized scores. By doing so, we hope to contribute to current debate on the issues of algorithmic bias, transparency, and fairness in machine learning. 
\end{abstract}
{\bf keywords:} Person, feature vector, user embedding, GDPR, moral identity, digital identity, humanistic view, AI

\section{Introduction}

\begin{quote}
    \emph{Explication is elimination: We start with a concept the expression for which is somehow troublesome; but it serves certain ends that cannot be given up. } {- John Rawls}
\end{quote}

Personalized products and services are an inescapable fact of modern life. From personalized predictions (Uber ETA, recidivism risk scores, credit scores), to personalized recommendations (Netflix, Spotify, Amazon, Tinder), personalized treatments (medicine and psychiatry), personalized ads, offers, prices, and more, we are inundated with personalized scores\footnote{We refer to recommendations, predictions, and treatments more generally as "scores."} based on data about our physical condition, genetic makeup, location, measurable behaviors and interactions. Ideally, personalization reduces information overload and search costs, thereby improving decision making and user experience. The decision maker and scored individual can be the same (self-directed) as in recommender systems, or different (other-directed) as in decision support systems used in law, medicine, business, and finance. In both cases, 
personalization relies on machine learning (ML). 

Nevertheless, ML personalization remains an ambiguous and under-examined concept. A
literature search reveals a surprising lack of consensus on its essential characteristics. 
It is either not explicitly defined, explained circularly, or used in relation to or interchangeably with \emph{customization}, \emph{tailoring}, or \emph{precision marketing}.
\citet{cremonesi2010performance} say what personalization is \emph{not}: giving "any user a pre-defined, fixed list of items, regardless of his/her preferences." Likewise, in the paper
"What Is Personalization?" \citet{fan2006personalization} classify various approaches to personalization. 
Despite offering a useful taxonomy of ideal personalization types, 
the authors overlooked a simple, but fundamental point: the user is a \textit{person}.
Thus, critical questions arise: Does merely assigning a unique score to each user constitute a personalized score? Is a score personalized when it is based (wholly or in part) on the behavior of other persons? Does it matter \emph{who} these "others" are? Does personalization require \emph{personal data}?

Such questions are critical because lawmakers, journalists, social scientists, managers, and the general public encounter ML personalization everywhere, yet each group may have a very different understanding of the process.  These differences can lead to confusion about the origin and accuracy of so-called "personalized" scores. This is especially relevant in light of recent regulations, such as the General Data Protection Regulation's (GDPR) \emph{right to explanation} for cases of automated profiling and its inclusion of a \emph{right to be forgotten}.

 This paper contrasts the process of ML personalization with the humanistic concept of a person and offers some normative criteria for evaluating the level of personalization. 
A clearer understanding of ML personalization will also contribute to current debates on algorithmic bias, transparency, and fairness in machine learning.

\section{The Machine Learning Personalization Perspective: The Person as a Feature Vector}
\label{sec-ml}

\begin{figure}[h]
    \centering
    \includegraphics[width=1\textwidth]{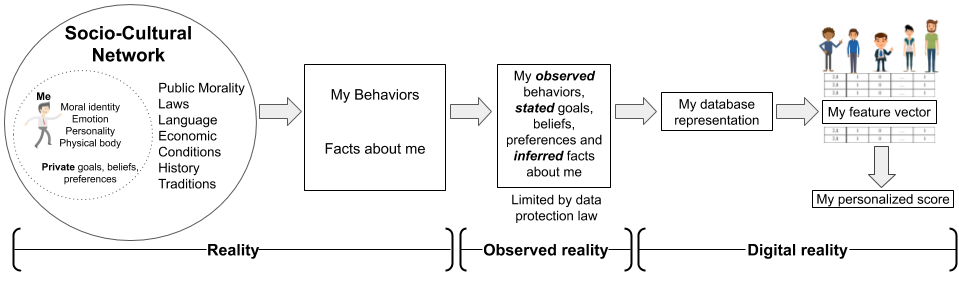}
    \caption{From humanistic concepts of the person to personalized scores: ML operationalization requires narrowing a person embedded in social and cultural space to a feature vector embedded in feature space.}
    \label{fig:schematic}
\end{figure}

Most commercial personalization engines (e.g. recommender systems) rely on a combination of explicitly-given preferences, demographic data, and observed behavior assumed to reflect one's underlying preferences, (i.e., "implicit ratings") such as purchases, repeated use, save/print, delete, reply, mark, glimpse, query \citep{nichols1998implicit}. Such data are derived only from what is \emph{actually} observed when using an application or device and are a small subset of our \emph{possibly} observed behaviors. 
Further, this narrow subset of recorded behaviors must be converted into a digital (database) representation supporting ML algorithms. Information is lost during the transformation of unstructured data (text, pictures, images, videos) into structured data which can be represented in matrix form. Lastly, the database representation is projected into \emph{feature space}. Figure \ref{fig:schematic} illustrates the ML pipeline resulting in 
the operationalization of a person as a feature vector.

A key idea in ML personalization rests on the metaphor of the \emph{person as a feature vector}. A feature vector (also known as a \emph{user embedding}) is an array of numbers--each representing some descriptive feature or aspect of an object--which can be thought of as the axes of a coordinate system \citep{kelleher2015fundamentals}. A 10-dimensional feature vector of a person, for instance, represents a person as an array of 10 numbers, derived from measurements of their observed behavior, and replaces the "person" with a single point in 10-dimensional feature space. Once converted to a point in feature space, the "similarity" of this person to others can be computed by measuring the distance between this point and others in the feature space. Points closer to each other are deemed more similar. Depending on one's modeling and predictive goals, this can indeed be a useful approach. But like any good metaphor, the feature vector approach emphasizes certain similarities at the expense of certain dissimilarities. When we represent a person as a feature vector, we tend to forget the long abstraction process that preceded it. As just one example of a potential problem in this process,  \citet{barocas2017fairness} remark, "[numeric] features incorporate normative and subjective \ldots assumptions about measurements, but all features are treated as numerical truth by  ML systems." In short, three key characteristics of ML personalization arise from the above process: ML personalization  1) is highly behaviorist in its assumptions, 2) is extremely narrow in predictive scope, and 3) often relies on data of a "community" whose selection is based on predictive goals. 

\subsection*{Three Characteristics of ML Personalization: Behaviorism, Predictive Scope, and Community}
\vspace{-0.1in}

First, \emph{ML personalization is strongly behaviorist}. In the behaviorist worldview, "reality" is only what can be observed \citep{van2013machine}. Behaviorism seeks to eliminate the messy theoretical causal mechanisms of human behavior (beliefs, intentions, goals, etc.) and focus instead on what can be measured and recorded \citep{skinner1965science}. ML personalization takes this idea one step further by focusing on predictive accuracy rather than on 
causal factors leading to the observed behavior \citep{rudin2019stop}. 

In the representation of the person as a feature vector, "important" attributes are only those that contribute to the algorithm's 
predictive accuracy. The predictive goal of ML personalization leads to valuing certain aspects of human behavior more than others; for example, behaviors most amenable to measurement tend to be recorded. This leads to a form of selection bias in the representation of the person. Unpredictable behavior is deemed noise, and measured behaviors which do not add to predictive accuracy are deemed redundant. Extraneous features can lead to the "curse of dimensionality" and 
are removed in order to preserve model parsimony,  storage space, and computational efficiency. One example of how ML personalization favors parsimony is the use of unsupervised dimension reduction techniques, such as Singular Value Decomposition (SVD). These techniques rely on the discovery that many of our measured behaviors are simply linear combinations of other behaviors and can thus be discarded with little information loss. We will see how issues of parsimony, scale, and computational efficiency constrain the degree of personalization when we examine Instagram's Explore, which must select relevant content for a focal user from hundreds of millions of other potential users' contributions. 

As we discuss in Section \ref{sec-humanistic}, the ML approach is at odds with a humanistic view of the person conceived as having an inner and outer self. As persons, our self-conceptions are highly bound up with our reasons for action, and these reasons may stem from our identity.  If the reasons we give for our actions are indeed \emph{causes} of our action, as many philosophers believe \citep[see, e.g.,][]{davidson1963actions}, then we lose an important aspect of the human experience by focusing solely on prediction. The representation of a person as a feature vector abstracts a "person" away from her socio-cultural network of meanings, identities, and values and represents her as a precisely-defined digital object in order to predict a very specific behavior, limited to a specific application (see Figure \ref{fig:schematic}). Human societies evolve over time and vary in their interpretations of codes of conduct regulating behavior (i.e., morality); yet, due to their abstract generality, mathematical objects and numbers do not. Further, the definitions and properties of mathematical objects remain fixed as a \emph{formal} public system of rules, whereas human morality is an \emph{informal} public system of rules of conduct in which the meaning of behavior is often unclear and evolving \citep{gert1998morality}. This is one possible source of the disconnect in the feature vector representation of the person used in ML personalization.

Second, \emph{ML personalization focuses on predicting a very narrow set of possible behaviors, often limited by the context of the application}. This can make it seem more powerful and accurate than it really is, especially when predictive performance is evaluated. For example, \citet{yeomans2019making} compared the predictions by friends and spouses to those from a simple collaborative filtering (CF) system for predicting a focal user's ratings of jokes. The study found the basic CF system was able to predict more accurately than a friend or spouse. Yet, the experiment included a set of 12 jokes \emph{pre-selected} by the researchers. The much more difficult problem of selecting 12 jokes from a nearly infinite set of possible jokes in all cultures and languages was left to the humans. In essence, the researchers had already personalized a list of jokes to each subject in the study, given their linguistic background, country of origin, and current location. Once narrowed to such a small recommendation space, the algorithm's performance appears quite impressive, but nevertheless hides the fact the hardest task had already been done by humans. A similar argument can be made for personalization on e-commerce sites: by going to a website, a person has already self-selected into a group 
who would be interested in products offered by the website. 
 Lastly, recidivism prediction in the judicial setting provides another example of the narrow predictive goal of ML personalization. Small changes in the definition of recidivism and its time-horizon-- Is it one, two, or three years in the future? Does it include misdemeanors or only felonies?--will drastically affect the algorithm's predicted scores and performance. 


Third, \emph{ML personalization uses data not only from the focal user, but also from other users}. This is clear in the case of recommendation systems using social network data, where the interests and preferences of a focal user can be inferred from their direct and indirect connections to other users and groups. But this point is less obvious when when the data are from measured behaviors on an application or device. Though recommendations based on data from other users could lead one to believe that one's personalized scores are not, in fact, "personal," in the sense of stemming from one and the same person, the real issue is whether the "community" chosen reflects one's \emph{chosen} social identity. In other words, personalization can still occur when the personalized scores are based on the behavior of others sharing the same social identity as the focal user. However, this is unlikely to occur in many ML contexts because "nearest neighbor" threshold values are chosen in order to optimize a given error metric, such as RMSE or precision/recall/top-N, not on the basis of any shared social identity \citep{cremonesi2010performance}. Recommendations based on the behavior of others who do not share a similar social and moral identity with the focal user are less personalized under our conception of personalization--all things equal. This principle also implies that when dimension reduction techniques are used, such as PCA and SVD (which rely on calculations of total variance across the entire user-item matrix), the selection of a "community" of users making up the user-item matrix is important. Personalized scores using latent factors derived from users who do not share a similar social or moral identity  are less personalized in our view. Of course, determining who is truly "similar" to us is a difficult question, but it is one where the methods and concepts of the humanities and social sciences may play an important role. 

 \subsection*{ML Personalization Case Study: Instagram's Explore}
 \vspace{-0.1in}

The personalization system of the popular social network Instagram shapes the desires, preferences, and self-conceptions of millions of people every day. By determining which content is displayed to which users, social media personalization systems may play an important role in the formation and maintenance of one's online and offline identity \citep{helmond2010identity}. When scaled to hundreds of millions of users, ML personalization can even have society-wide effects, shifting opinions towards political issues and candidates, as in the 2016 US presidential election. For this reason, we believe it is important to see how ML personalization relates to the "person" and its tacit assumptions regarding observed behavior. We therefore briefly describe Instagram's Explore to illustrate the person as a feature vector approach and the implications discussed in the previous section.  

The complicated set of algorithms used for Instagram's Explore recommendations essentially work by creating "account embeddings" in order to infer topical similarity.\footnote{Example taken from \url{ai.facebook.com/blog/powered-by-ai-instagrams-explore-recommender-system/}} 
Roughly put, a feature vector (FV) representation of an Instagram user is derived from the sequence of account IDs a user has visited. Next, the FV is used to determine a "community" of similar accounts.  
Similarity in this space implies topical similarity,  assuming users do not randomly explore different topics, but instead tend to view multiple accounts offering similar content.  Once the candidate accounts (the nearest neighbors) have been found, a random sample of 500 pieces of eligible content (videos, stories, pictures, etc.) is taken and a "first-pass" ranking is made to select the 150 "most relevant" pieces of content for a focal user. Then a 
neural network narrows the 150 to the top 50 most relevant. Notice that at each stage, possible content is narrowed-down and that computational efficiency is a high priority. Finally, a deep neural network is used to predict \emph{specific actions} (narrowed only to those possible on the Instagram app) for 25 pieces of content, such as "like," "save," or "see fewer posts like this." Content is then ranked based on a weighted sum of these very specific behaviors permitted in the app and their associated predicted probabilities. 
Weights are not determined by users; they are determined by the system engineers and reflect their assumptions about the users' intentions. Assuming a user wishes to discover new topics and avoid seeing content from the same users, a simple heuristic rule is used to downrank certain content and diversify the results. After applying the downranking procedure, the content with the highest weighted sum in the "value model" is displayed in decreasing order on the focal user's Explore page.  


\section{The Person, Personal Data, and the Legal Issues of Personalized Scores}
As described and illustrated in Section \ref{sec-ml}, personalized scores are generated by combining information about your observed behavior with certain assumptions about you and your goals, beliefs, and preferences to predict a very constrained set of future actions. But \emph{which information} should constitute me as a person, and from \emph{whose viewpoint}? These are difficult philosophical questions. 
To answer we might first ask, "What is a person?" But this question is problematic because "person" is often defined in such a way as to justify a particular world-view. It can variously mean "human animal,” "moral agent,” "rational, self-conscious subject”, "possessor of particular rights,” and "being with a defined personality or character"  \citep{schechtman2018constitution}. Nevertheless, Western thought has focused on two general aspects to the person, one of which is revealed through the etymology of the word itself. 

"Person" comes from the Latin \textit{persona}, which derives from the Ancient Greek word for a type of mask worn by dramatic actors. For the Ancient Greeks, the idea of a person was inherently connected to context and role. One's persona was a specific kind of self-identity that was public, socially defined, and varied depending on context. In time, the outward-facing Greek conception was complemented by a Christian emphasis on self-reflection and awareness, resulting in a focus on the human capacity for rational introspection \citep{douglas1998missing}. Later thinkers expanded the idea of the person into two main parts: an outer \textit{personnage} of public roles and masks, and an inner conscience, identity, and consciousness.  In short, Western thought has generally viewed the person as consisting in two co-existing inner and outer domains. 

We note that the word personalization contains the adjective \emph{personal}, implying that personalized scores should, at least in part, be based on \emph{personal data}. For example, \citet{liang2006personalized} write that personalization is "a process of collecting and using \textit{personal} information to uniquely tailor products, content and services to an individual." Yet, researchers in ML have largely overlooked the crucial connection between personalized scores and legal definitions of personal data.
A source of complexity is that different legal regimes define personal data differently.
For instance, the most influential data protection law at the moment, the EU’s GDPR, defines personal data as "any information relating to an identified or identifiable natural person (`data subject')" (Article 4). 
Such a broad definition means personal data could constitute anything from browser cookies, to location data, to even a combination of non-sensitive measurements, if there are sufficiently few or unique observations to single out individuals. Ultimately, context determines whether data are personal data \citep{greene2019adjusting}. 

Consent is another legal issue in the generation of personalized scores, especially when \emph{sensitive data} such as race, gender, or political affiliation are involved. The GDPR's reliance on explicit consent has important moral implications. Consent implies choice in how one represents oneself publicly and also which inferences data processors can draw about one's identity \citep[see, e.g.,][]{kosinski2013private}. The GDPR's right to rectification and erasure (the so-called "right to be forgotten") further entrenches the importance of autonomy in deciding the factual basis on which one will be judged, classified, and potentially discriminated against, and the extent to which our identities as persons are reflected in observations of our public behavior. This notion has been called \emph{informational privacy} \citep{shoemaker2010self, moreau2010discrimination}.
In short, the existing literature on ML personalization overlooks two key points: first, that it depends on differing legal definitions of personal data; and second, informational privacy concerns are generally not considered in the design of personalized systems. 

\section{Identity, Self, and Data Regulation}
\label{sec-humanistic}

ML researchers and practitioners normally deal with issues of “practical identity,” while philosophers and social scientists deal with issues of "moral" and "social" identity \citep{manders2010practical, de2010identity}. Recently, humanistic scholars have focused on a new, third kind of identity arising from the difficulties in the formation of a single, unitary digital identity in light of rapidly evolving technology. The nature of distributed storage and collection of observable behavioral data related to living persons further complicates the process of identity formation (see "Digital reality" in Figure \ref{fig:schematic}). These distributed digital representations of our practical, moral, and social identities have been variously termed \emph{data citizens},  \emph{data doubles}, \emph{data shadows} and \emph{digital subjects} \citep{goriunova2019digital}.  

The narrowing from reality, to observed reality, to digital reality (Fig 1) raises questions about the relation between persons, personal data, and personalized scores. 
While many of today's ML algorithms generate different scores given different contextual input data, these scores can fail to reflect our\emph{ moral identities}. For example, scores based on categorical features may not reflect one's social or personal identity. Similarly, explanations for personalized scores at the algorithm or model level (e.g., variable importance scores) may fail to reflect the salience of beliefs, motivations, and intentions we have when explaining our behavior to others. An explanation of a personalized score that leaves out these important aspects of one's moral identity surely cannot be said to relate to one \textit{as a person}. Though we cannot control how society sees us, we can choose to (partly) accept or reject society’s categorization of us. This latter aspect of human experience remains largely unaccounted for in ML personalization. As \citet{manders2010practical}  argues, "The challenge [of] justice to data subjects as moral persons is to take into account the self-informative perspective that is part of `identity management.'" 


We now provide a brief overview of identity theories and humanistic perspectives 
drawing mainly from social and personality psychology, consumer behavior, sociology, philosophy, economics, information systems, and recent data protection law. We hope exposure to these ideas can influence future thinking on and design of ML systems generating personalized scores.
Table \ref{tab:perpectives} summarizes key concepts 
related to the person from these 
disciplines. 

\subsubsection*{Identity from a Social Science Perspective} \label{subsec-psy-soc}
\vspace{-0.1in}

Theories and concepts of self and identity are fundamental topics in psychology, despite being difficult to precisely define. Personality psychologists \citet{larsen2009personality} define personality as a an organized set of relatively enduring psychological traits that guide a person's interactions with her "intrapsychic, physical, and social environments." These traits can be expressed as the "Big Five" dimensions of personality:    Openness to New Experience, Conscientiousness, Extraversion, Agreeableness, and Neuroticism (also known as OCEAN), and have been shown to have varying levels of predictive power.\footnote{
\citet{mischel2007introduction} show that knowing an individual's Big Five trait profile allows for only weak prediction of a particular behavior in a particular context} 
Personality psychology has also begun to study the formation of \textit{moral identity}. For example, \citet{aquino2002self} show how highly important moral identities--the collection of certain beliefs, attitudes, and behaviors relating to what is right or wrong--can provide a basis for the construction of one's "self-definition." Finally, social psychologists define the self as an interface between the biological processes of the body and the larger social and cultural context.  Further, they highlight the power of the cultural context in defining identity: "Without society, the self would not exist in full"  \citep[p. 74]{baumeister2014social}. As the summaries above illustrate, understanding the person requires description at multiple levels.  \citet{mcadams1995we} presents an influential multi-dimensional theory of the person, focusing on traits, personal concerns, and life stories.

According to identity theorists in sociology and psychology, the self is fluid and occupies multiple social roles (identity theory) or group identities (social identity theory) that coexist and vary over time. The dynamism of the self-concept is reflected in its numerous sub-components or self-representations: the past, present, and future self; the ideal, "ought," actual, possible, and undesired self \citep{markus1987dynamic}.  
Identity theorists highlight how a unified self-conception arises from the variety of meanings given to various social roles the self occupies \citep{stryker2000past}. Common types of social identities relate to one's ethnicity, religion, political affiliation, job, and relationships. People may also identify strongly with their gender, sexual orientation, and various other "stigmatized" identities, such as being homeless, an alcoholic, or overweight (Deux 2001). In short, social psychologists generally agree one's self-concept is one of the most important regulators of one's behavior.

The self-concept has also influenced postmodernist consumer behavior research. Consumers now take for granted their sometimes paradoxical identities, beliefs, and behaviors in everyday life \citep{fuat1995marketing}. In the postmodern condition one "listens to reggae, watches a Western, eats McDonald's food for lunch and local cuisine for dinner, wears Paris perfume in Tokyo and retro clothes in Hong Kong" \citep{lyotard1984postmodern}. 

\subsubsection*{Identity, the Self, and Morality from a Philosophical Perspective}\label{subsec-philosophy}
\vspace{-0.1in}
How do philosophers understand the link between the identity and the self? For some, the self arises from an identification of structural sameness over time, experienced as a kind of narrative \citep{ricoeur1994oneself}. Though individual cells of our bodies are constantly renewed, a person's essential structure nevertheless remains similar enough to be identified as the "same" over time. For others, such as \citet{taylor1989sources}, social and communal life link the two concepts. \citet{taylor1989sources} argues that one's identity is tied to one's moral values and a defining social community. He writes, "I define who I am by defining where I speak from, in the family tree, in social space, in the geography of social statuses and functions, in my intimate relations to the ones I love, and also crucially in the space of moral and spiritual orientation within which my most important defining relations are lived out." Our identities  therefore rest on the constant interplay between how others view us and how we view ourselves. These "community" identities further entail various moral obligations to others 
that influence and motivate our behavior .

 \citet{atkins2010narrative} draws on work by Ricoeur and Korsgaard to explain the connection between morality and identity. She claims that "one's identity is the source of one's moral agency, expressed in one's normative reasons." One's identity constitutes "the condition of the possibility for having a perspective from which to perceive, deliberate, and act \ldots it is a condition of the possibility of morality." For both Taylor and Atkins, the development of a self-identity provides a moral scaffolding upon which possible actions can be evaluated; our identities are critical for embedding us both in the physical and social worlds, and adding a moral dimension to our actions as persons. 

 In  \citet{schechtman2018constitution}'s influential narrative self-constitution view, personhood and personal identity are fleshed out in terms of a unified and "unfolding developmental structure,” analogous to a complex musical sonata. Echoing the ideas of Ricoeur,  she writes that human life "is a structural whole that has, by its very nature, attributes that apply to it as a whole which do not necessarily apply to each individual portion." Through the unity of experience--which takes the form of a narrative--the \textit{same person} can be an infant at one time and later in life suffer from dementia. Her notion of personal identity is similar to McAdams’ psychological view in that it explains via a diachronic process how identity can evolve over time. These views highlight the processual, dynamic, and emergent nature of the self. 

\subsubsection*{Identity in Information Systems (IS) and Economics}
\vspace{-0.1in}

As the World Wide Web grew in popularity, scholars in the field of IS began to examine its impact not just on the organization, but on the individual. Some were influenced by postmodern arguments and pushed for a reconsideration of such fundamental concepts as space, time, the world, and the self \citep{introna1996information}. \citet{erickson1996world}, for example, foreshadowed the rise of social media when he examined the "portrayal management" of online identities afforded by personal webpages. The anonymity of the Internet meant that individuals could selectively represent themselves to others without the complexities of physical contact. Nearly two decades later, the study of online communities and identity formation has entered into the mainstream, leading to novel methods of data collection and analysis. 
A growing stream of research in IS now draws on the constructs of identity and socio-cultural norms to explain online behavior. 
\citet{burtch2016secret}, for example,  explored how contributing to the same crowdfunding campaign could foster a common identity in users and influence a campaign's fundraising success. Finally, in the more traditional domain of economics, \citet{akerlof2000economics} presented one of the first and most influential economic analyses of the role of social and group identity on behavior. Their work illustrates how decisions about identity can be used to explain apparently irrational behavior, the creation of behavioral externalities in others, and preference evolution.

\begin{table}
    \centering
    \caption{Perspectives and Key Concepts on the Person and Personal Identity}
    \label{tab:perpectives}
    \begin{footnotesize}
    \begin{tabular}{|p{0.22\linewidth}|p{0.27\linewidth}|p{0.22\linewidth}|p{0.23\linewidth}|}
      \toprule
    {\bf Philosophy \& Humanities}  
    & {\bf Psychology, Sociology \& Consumer Behavior} 
    & {\bf Economics  \& Information Systems} 
    & {\bf Data Regulation \& Professional Codes of Conduct}  \\ \midrule
    Moral identity & Personality: Big 5 & Identity explains irrational & Right to be forgotten\\
    Community-defined identity & Socially-defined identity & behavior & Right to rectification \\
    Pursuit of the good life & Culturally-influenced narratives & Digital identity & Automated profiling \\
    The Postmodern experience & Emotions and moods & Online identity affects & Informational privacy \\
    Narrative identity & Hermeneutic interpretation & behavior & Agency over digital identity \\
    Inner and outer-facing self & Meaning of symbols, texts, images & Identity as economic choice & Ethically-aligned design    \\    \bottomrule
    \end{tabular}

    \end{footnotesize}
\end{table}

\subsubsection*{Digital Identity in Data Regulation and Professional Codes of Conduct}
\vspace{-0.1in}

 Around the world, the passage of new data protection laws reflects growing social concern over the process of turning a person into a feature vector. The GDPR and the California Consumer Privacy Act (CCPA) both seek to give citizens more control over their personal data and their digital identities. The GDPR gives data subjects rights to opt out of "automated profiling" with "legal" or other "significant" effects.   The GDPR's \emph{right to be forgotten} can be viewed as a potential legal remedy to the issue of one's practical and moral identity diverging over time. Data subjects have the right to destroy and recreate aspects of their digital identities when they no longer represent them as persons. At the same time, important ML societies, such as the IEEE and ACM, have also voiced concern about the effects of ML on human well-being. For example, recent IEEE guidelines state that autonomous and intelligent systems should give people access and control over their personal data and allow them "agency over their digital identity."\footnote{The IEEE Global Initiative on Ethics of Autonomous and Intelligent Systems. \emph{Ethically Aligned Design: A Vision for Prioritizing Human Well-being with Autonomous and Intelligent Systems}, 1st ed, 2019 \url{https://ethicsinaction.ieee.org}}  
The following section thus lays out a basic conceptual framework for use by regulators, professional groups, and practitioners to assess levels of personalization.

\section{Evaluating ML Personalization}

In order to better align ML personalization with a humanistic view of the person and current trends in data protection law, we propose six criteria to guide future discussions. 
We consider the degree of personalization to be on a continuum and to depend on 
objective (outer-facing) and subjective (inner-facing) dimensions. As shown earlier in Figure \ref{fig:schematic}, the inner aspects of the person are inaccessible through observational methods. First-person, subjective input from the person herself is therefore crucial in measuring the level of personalization. The social sciences and humanities are especially well-positioned to contribute to the interdisciplinary development of such measures using quantitative and qualitative 
data collection and analysis methods. 
Lastly, legal experts will also be needed to relate these measures to data subjects' rights under current data protection laws, such as the GDPR.  

{\bf Objective dimensions} include (1) a \emph{personal data component} - the extent to which legally-defined personal data are used, and (2) a \emph{uniqueness component} - the percent of other users sharing the same input data, personalized score, or discretized recommendation. The personal data component implies that legally-allowable degrees of personalization may vary by country and legal regime. Regarding the uniqueness component, it is not clear how to weight these three aspects in determining the degree of personalization. Numeric measurements may uniquely identify you more so than categorical measurements, but these categories may more closely reflect your personal identity and community. Similarly, it is not obvious how to reconcile
dimension reduction techniques with personalization. For example, when highly morally-salient measurements are correlated to other measures, should they be removed?  

Another objective dimension worth examining is (3) an \emph{accuracy component} - the accuracy of predicted scores. It seems reasonable to assume  personalized predictions should be more accurate than non-personalized predictions. But how should accuracy be measured?
 As we noted above, the very narrow predictive goal of ML personalization can obscure the interpretation of some accuracy metrics. An easy way to improve accuracy is simply to narrow the size of the "recommendation set."
However, in some cases, accuracy may not be desired; some persons may prefer to trade off predictive accuracy (in the narrow sense above) for informational privacy. As noted earlier, data subjects have basic rights to informational privacy under the GDPR. 
 

{\bf Subjective dimensions} include (4) a \emph{self-determination component} - the extent to which personalized scores respect the data subject's right to determine his public self, captured in the notion of informational privacy. This subjective component is relevant to debates on fair and transparent ML (FATML). Current legal and ML approaches to fairness operate on the level of the "database representation" of the person (i.e., practical identity), overlooking the person's moral identity. Thus we propose a (5) \textit{right reasons component} - fair decisions should capture the person's intended public self and associated goals, attitudes, meanings and motivations in that context.  The relevance of reasons given in algorithmic explanations can be assessed, for example, using relevance feedback techniques already in use in information retrieval. Lastly, another subjective dimension is (6) a \emph{moral importance component} - the moral importance of the personalization context. 
For example, how should the different subjective and objective components be weighted when assessing a system that recommends a movie versus a bail amount? 


Finally, our proposed list of evaluation components is by no means exhaustive. In fact, some of these components might interact with or even contradict other aspects of personalization. For example, 
serendipitous recommendations\footnote{Serendipity is used in Instagram's "downranking" procedure, under the assumption that one's preferences should not remain static and so new topics should be presented in order to provide diverse personalized content.}  \citep{ge2010beyond} can stimulate evolution in one's moral identity, yet might conflict with the \emph{accuracy component}. Allowing users to decide for themselves how they would like to weight these aspects may be one solution.

\section{Future Work and Conclusion}
 \vspace{-0.1in}
 Going forward, we believe that as ML personalization becomes increasingly rooted in daily life, we should consider new approaches to data collection, such as those used in qualitative social science (e.g., hermeneutics, discourse analysis, fuzzy-set theory). By incorporating more diverse forms of personal data into personalized scores, we may be able to reduce the gap between one's identity as a \textit{person} embedded in social and cultural space and as a \textit{feature vector} embedded in feature space. As we emphasize, however, the ML personalization process will never be perfect. 

Our presentation of the ML pipeline can also be expanded to consider further ML operations, such as how and which data are used for training algorithms. For example, personalized scores for a given person are computed using models that were trained on data from other persons, typically excluding that person's own data. Chronology of the training set is also important when considering the dynamism of a person's self-concept. 

Lastly, these evaluation dimensions are merely an entry point into an interdisciplinary discussion about personalization. 
We hope to initiate and contribute to such a discussion by describing what the process of ML personalization typically looks like, particularly its defining metaphor of the person as feature vector. We argued that this conception is radically different from that found in the humanities, social sciences, and law. As the long-term effects of ML personalization on personal identity, politics, law and society are still unclear, it is important to critically examine the process of ML personalization. Perhaps it cannot be precisely defined, but an examination of its key characteristics and assumptions may help foster new insights on the issues of algorithmic bias, transparency, and fairness. 

\section*{Acknowledgements}\vspace{-0.1in}
We thank Ching-Fu Lin, Jack Buchanan, Mariangela Guidolin, Kellan Nguyen, and Boaz Shmueli for their helpful comments on an earlier draft of this paper. 

\bibliographystyle{apa}  
\bibliography{personalization.bib}

\begin{thebibliography}{}

\bibitem[\protect\astroncite{Akerlof and Kranton}{2000}]{akerlof2000economics}
Akerlof, G.~A. and Kranton, R.~E. (2000).
\newblock Economics and identity.
\newblock {\em Quarterly Journal of Economics}, 115(3):715--753.

\bibitem[\protect\astroncite{Aquino et~al.}{2002}]{aquino2002self}
Aquino, K., Reed, I., et~al. (2002).
\newblock The self-importance of moral identity.
\newblock {\em J of personality and soc psych}, 83(6):1423.

\bibitem[\protect\astroncite{Atkins}{2010}]{atkins2010narrative}
Atkins, K. (2010).
\newblock {\em Narrative identity and moral identity}.
\newblock Taylor \& Francis.

\bibitem[\protect\astroncite{Barocas et~al.}{2017}]{barocas2017fairness}
Barocas, S., Hardt, M., and Narayanan, A. (2017).
\newblock Fairness in machine learning.
\newblock {\em NIPS Tutorial}.

\bibitem[\protect\astroncite{Baumeister and
  Bushman}{2014}]{baumeister2014social}
Baumeister, R.~F. and Bushman, B. (2014).
\newblock {\em Social psychology and human nature}.
\newblock Belmont, CA: Cengage Learning.

\bibitem[\protect\astroncite{Burtch et~al.}{2016}]{burtch2016secret}
Burtch, G., Ghose, A., and Wattal, S. (2016).
\newblock Secret admirers: An empirical examination of information hiding and
  contribution dynamics in online crowdfunding.
\newblock {\em Information Systems Research}, 27(3):478--496.

\bibitem[\protect\astroncite{Cremonesi et~al.}{2010}]{cremonesi2010performance}
Cremonesi, P., Koren, Y., and Turrin, R. (2010).
\newblock Performance of recommender algorithms on top-n recommendation tasks.
\newblock In {\em Proceedings of the fourth ACM conference on Recommender
  systems}. ACM.

\bibitem[\protect\astroncite{Davidson}{1963}]{davidson1963actions}
Davidson, D. (1963).
\newblock Actions, reasons, and causes.
\newblock {\em The journal of philosophy}, 60(23):685--700.

\bibitem[\protect\astroncite{De~Vries}{2010}]{de2010identity}
De~Vries, K. (2010).
\newblock Identity, profiling algorithms and a world of ambient intelligence.
\newblock {\em Ethics and information technology}, 12(1):71--85.

\bibitem[\protect\astroncite{Douglas and Ney}{1998}]{douglas1998missing}
Douglas, M. and Ney, S. (1998).
\newblock {\em Missing persons: A critique of the personhood in the social
  sciences}, volume~1.
\newblock Univ of California Press.

\bibitem[\protect\astroncite{Erickson}{1996}]{erickson1996world}
Erickson, T. (1996).
\newblock The world-wide-web as social hypertext.
\newblock {\em Communications of the ACM}, 39(1):15--17.

\bibitem[\protect\astroncite{Fan and Poole}{2006}]{fan2006personalization}
Fan, H. and Poole, M.~S. (2006).
\newblock What is personalization? perspectives on the design and
  implementation of personalization in information systems.
\newblock {\em Journal of Org. Comp. and Elec. Comm.}, 16(3-4):179--202.

\bibitem[\protect\astroncite{Fuat~Firat et~al.}{1995}]{fuat1995marketing}
Fuat~Firat, A., Dholakia, N., and Venkatesh, A. (1995).
\newblock Marketing in a postmodern world.
\newblock {\em European journal of marketing}, 29(1):40--56.

\bibitem[\protect\astroncite{Ge et~al.}{2010}]{ge2010beyond}
Ge, M., Delgado-Battenfeld, C., and Jannach, D. (2010).
\newblock Beyond accuracy: evaluating recommender systems by coverage and
  serendipity.
\newblock In {\em Proceedings of the fourth ACM conference on Recommender
  systems}. ACM.

\bibitem[\protect\astroncite{Gert}{1998}]{gert1998morality}
Gert, B. (1998).
\newblock {\em Morality: Its nature and justification}.
\newblock Oxford University Press on Demand.

\bibitem[\protect\astroncite{Goriunova}{2019}]{goriunova2019digital}
Goriunova, O. (2019).
\newblock The digital subject: People as data as persons.
\newblock {\em Theory, Culture \& Society}, 36(6):125–145.

\bibitem[\protect\astroncite{Greene et~al.}{2019}]{greene2019adjusting}
Greene, T., Shmueli, G., Ray, S., and Fell, J. (2019).
\newblock Adjusting to the gdpr: The impact on data scientists and behavioral
  researchers.
\newblock {\em Big data}, 7.

\bibitem[\protect\astroncite{Helmond}{2010}]{helmond2010identity}
Helmond, A. (2010).
\newblock Identity 2.0: Constructing identity with cultural software.
\newblock In {\em Proceeding of Mini-conference initiative, University of
  Amsterdam (Amsterdam, ND--Jan 20-22 2010)}. Citeseer.

\bibitem[\protect\astroncite{Introna and
  Whitley}{1996}]{introna1996information}
Introna, L.~D. and Whitley, E.~A. (1996).
\newblock Information systems as a social science? the individual perspective.
\newblock {\em Information Systems}, 8:16--1996.

\bibitem[\protect\astroncite{Kelleher et~al.}{2015}]{kelleher2015fundamentals}
Kelleher, J.~D., Mac~Namee, B., and D'arcy, A. (2015).
\newblock {\em Fundamentals of machine learning for predictive data analytics:
  algorithms, worked examples, and case studies}.
\newblock MIT Press.

\bibitem[\protect\astroncite{Kosinski et~al.}{2013}]{kosinski2013private}
Kosinski, M., Stillwell, D., and Graepel, T. (2013).
\newblock Private traits and attributes are predictable from digital records of
  human behavior.
\newblock {\em Proceedings of the National Academy of Sciences},
  110(15):5802--5805.

\bibitem[\protect\astroncite{Larsen and Buss}{2009}]{larsen2009personality}
Larsen, R. and Buss, D.~M. (2009).
\newblock {\em Personality psychology}.
\newblock McGraw-Hill Publishing.

\bibitem[\protect\astroncite{Liang et~al.}{2006}]{liang2006personalized}
Liang, T.-P., Lai, H.-J., and Ku, Y.-C. (2006).
\newblock Personalized content recommendation and user satisfaction:
  Theoretical synthesis and empirical findings.
\newblock {\em Journal of Management Information Systems}, 23(3):45--70.

\bibitem[\protect\astroncite{Lyotard}{1984}]{lyotard1984postmodern}
Lyotard, J.-F. (1984).
\newblock {\em The postmodern condition: A report on knowledge}, volume~10.
\newblock U of Minnesota Press.

\bibitem[\protect\astroncite{Manders-Huits}{2010}]{manders2010practical}
Manders-Huits, N. (2010).
\newblock Practical versus moral identities in identity management.
\newblock {\em Ethics and information technology}, 12(1):43--55.

\bibitem[\protect\astroncite{Markus and Wurf}{1987}]{markus1987dynamic}
Markus, H. and Wurf, E. (1987).
\newblock The dynamic self-concept: A social psychological perspective.
\newblock {\em Annual review of psychology}, 38(1):299--337.

\bibitem[\protect\astroncite{McAdams}{1995}]{mcadams1995we}
McAdams, D.~P. (1995).
\newblock What do we know when we know a person?
\newblock {\em Journal of personality}, 63(3):365--396.

\bibitem[\protect\astroncite{Mischel et~al.}{2007}]{mischel2007introduction}
Mischel, W., Shoda, Y., and Ayduk, O. (2007).
\newblock {\em Introduction to personality: Toward an integrative science of
  the person}.
\newblock John Wiley \& Sons.

\bibitem[\protect\astroncite{Moreau}{2010}]{moreau2010discrimination}
Moreau, S. (2010).
\newblock What is discrimination?
\newblock {\em Philosophy \& Public Affairs}, 38(2):143--179.

\bibitem[\protect\astroncite{Nichols}{1998}]{nichols1998implicit}
Nichols, D. (1998).
\newblock Implicit rating and filtering.
\newblock In {\em Proc of 5th DELOS Workshop on Filtering and Collab
  Filtering}.

\bibitem[\protect\astroncite{Ricoeur}{1994}]{ricoeur1994oneself}
Ricoeur, P. (1994).
\newblock {\em Oneself as another}.
\newblock University of Chicago Press.

\bibitem[\protect\astroncite{Rudin}{2019}]{rudin2019stop}
Rudin, C. (2019).
\newblock Stop explaining black box machine learning models for high stakes
  decisions and use interpretable models instead.
\newblock {\em Nature Machine Intelligence}, 1(5):206.

\bibitem[\protect\astroncite{Schechtman}{2018}]{schechtman2018constitution}
Schechtman, M. (2018).
\newblock {\em The constitution of selves}.
\newblock Cornell university press.

\bibitem[\protect\astroncite{Shoemaker}{2010}]{shoemaker2010self}
Shoemaker, D.~W. (2010).
\newblock Self-exposure and exposure of the self: informational privacy and the
  presentation of identity.
\newblock {\em Ethics and Information Technology}, 12(1):3--15.

\bibitem[\protect\astroncite{Skinner}{1965}]{skinner1965science}
Skinner, B.~F. (1965).
\newblock {\em Science and human behavior}.
\newblock Simon and Schuster.

\bibitem[\protect\astroncite{Stryker and Burke}{2000}]{stryker2000past}
Stryker, S. and Burke, P.~J. (2000).
\newblock The past, present, and future of an identity theory.
\newblock {\em Social Psychology Quarterly}, 63(4):284--297.

\bibitem[\protect\astroncite{Taylor}{1989}]{taylor1989sources}
Taylor, C. (1989).
\newblock {\em Sources of the self: The making of the modern identity}.
\newblock Harvard University Press.

\bibitem[\protect\astroncite{Van~Otterlo}{2013}]{van2013machine}
Van~Otterlo, M. (2013).
\newblock A machine learning view on profiling.
\newblock {\em Privacy, Due Process and the Computational Turn-Philosophers of
  Law Meet Philosophers of Technology. Abingdon: Routledge}.

\bibitem[\protect\astroncite{Yeomans et~al.}{2019}]{yeomans2019making}
Yeomans, M., Shah, A., Mullainathan, S., and Kleinberg, J. (2019).
\newblock Making sense of recommendations.
\newblock {\em Journal of Behavioral Decision Making}, 32(4):403--414.

\end{thebibliography}
\end{document}